\documentclass{article}
\usepackage{arxiv}

\usepackage{graphicx}
\input{abbrv}

\usepackage[utf8]{inputenc} 
\usepackage{hyperref}
\usepackage{cleveref}
\usepackage{url}
\usepackage{wrapfig}
\usepackage{amsthm}
\usepackage{thmtools, thm-restate}
\usepackage{wasysym}
\usepackage{standalone}
\usepackage{array,multirow,booktabs}
\usepackage[square,numbers]{natbib}
\usepackage[inline]{enumitem}
\usepackage{xcolor}

\theoremstyle{definition}
\newtheorem{definition}{Definition}[section]
\newtheorem{theorem}{Theorem}[section]

\newcommand{\PR}{\text{Pr}}
\DeclareMathOperator*{\sound}{soundness}
\DeclareMathOperator*{\compl}{completeness}

\begin{document}
\title{Hardness of Deceptive Certificate Selection}
%
%
 \author{Stephan Wäldchen\orcidID{0000-0001-7629-7021}}
%
\author{%
  Stephan Wäldchen \\
 Zuse Institute Berlin \\
  \texttt{waeldchen@zib.de} \\
}

%
\maketitle              
\begin{abstract}
Recent progress towards theoretical interpretability guarantees for AI has been made with classifiers that are based on interactive proof systems. A prover selects a certificate from the datapoint and sends it to a verifier who decides the class. In the context of machine learning, such a certificate can be a feature that is informative of the class. For a setup with high soundness and completeness, the exchanged certificates must have a high mutual information with the true class of the datapoint. However, this guarantee relies on a bound on the Asymmetric Feature Correlation of the dataset, a property that so far is difficult to estimate for high-dimensional data. It was conjectured in~\cite{waldchen2022merlin} that it is computationally hard to exploit the AFC, which is what we prove here.

We consider a malicious prover-verifier duo that aims to exploit the AFC to achieve high completeness and soundness while using uninformative certificates. We show that this task is $\SNP$-hard and cannot be approximated better than $\mathcal{O}(m^{1/8 - \epsilon})$, where $m$ is the number of possible certificates, for $\epsilon>0$ under the Dense-vs-Random conjecture. This is some evidence that AFC should not prevent the use of interactive classification for real-world tasks, as it is computationally hard to be exploited.
\keywords{Interactive Proof Systems, Formal Interpretability, XAI, Merlin-Arthur classifiers}
\end{abstract}
\section{Introduction}

Safe deployment of Neural Network (NN) based AI systems in high-stakes applications, e.g., medical diagnostics~\cite{holzinger2017we} or autonomous vehicles~\cite{schraagen2020trusting},
requires that their reasoning be subject to human scrutiny.
One of the most prominent approaches in explainable AI (XAI) is feature-based interpretability, where a subset of the input values are highlighted as being important for the classifier decison in the form of saliency or relevance maps~\cite{mohseni2021multidisciplinary}.
Feature-based interpretability had some successes, such as detecting biases in established datasets~\cite{lapuschkin2019unmasking}. However, most approaches are motivated primarily by heuristics and come without any theoretical guarantees, which means their success cannot be verified. While they work well on many generic tasks, they fail when some effort is put into hiding the true reasoning process. It has also been demonstrated for numerous XAI-methods that they can be manipulated by a clever design of the NNs~\cite{slack2021counterfactual,slack2020fooling,anders2020fairwashing,dimanov2020you}.
This motives formal approaches for interpretability with clearly stated assumptions and guarantees that can be used by auditors to check the reasoning of commercial AI systems.

\subsection{Information Bounds on Features}

The most prominent approaches to define the importance of an input feature are Shapley Values~\cite{NIPS2017_7062shap}, Mutual Information~\cite{vergara2014review,chen2018learning} and Precision~\cite{ribeiro2018anchors}. All these concepts are closely related to each other, and the precision can be used to lower-bound the mutual information~\cite{waldchen2022merlin}. In principle, they all rely on the idea of partial input to the classifier $f:[0,1]^d \rightarrow \skl{-1.1}$~\cite{waldchen2022training}. For input $\bfx\in [0,1]^d$ and a set $S\subset[d]$, a partial input $\bfx_S$ consists of the values of $\bfx$ on the indices in $S$. Since most classifiers cannot evaluate partial input, it was proposed in~\cite{NIPS2017_7062shap} to consider the expectations over the input distribution $\CD$ conditioned on $\bfx_S$.
In the context of precision, this leads us to the definition 
  \[
   \PR_{f,\bfx}(S) := P_{\bfy\sim \CD}\ekl{f(\bfy)=f(\bfx)~|~ \bfy_S=\bfx_S} = P_{\bfy\sim \CD|_{\bfx_S}}\ekl{f(\bfy)=f(\bfx)}.
  \]
The challenge for these formal approaches is that $\CD|_{\bfx_S}$ is generally unknown. In fact, inexact modelling of this distribution is exactly what allowed for the manipulation of the methods in \cite{slack2021counterfactual,slack2020fooling,anders2020fairwashing,dimanov2020you}, as well as for the artefactual explanations produced by the RDE~\cite{macdonald2019rate} approach in some images~\cite{waldchen2022training}. While modelling this distribution for complicated data, e.g. images, can be practically achieved with deep generative networks~\cite{chattopadhyay2022interpretable}, there is no known way to bound how accurately this models the true distribution. To compare the modelled distribution to some ground truth would require a large amount of samples conditioned on many possible features $\bfx_S$, which is beyond the scope of even the largest datasets.  


As a remedy, interactive classification, inspired by Interactive Proof Systems, has been developed and demonstrated to yield lower bounds on the precision of features without having to calculate $\CD|_{\bfx_S}$ explicitely~\cite{waldchen2022merlin}.

\begin{figure}
    \centering
    \begin{tabular}{c@{\hspace{0.1cm}}|@{\hspace{0.1cm}}c}
    {\small Unsound Verifier}l & {\small Sound Verifier} \\[1em]
       \includegraphics[width=0.48\textwidth]{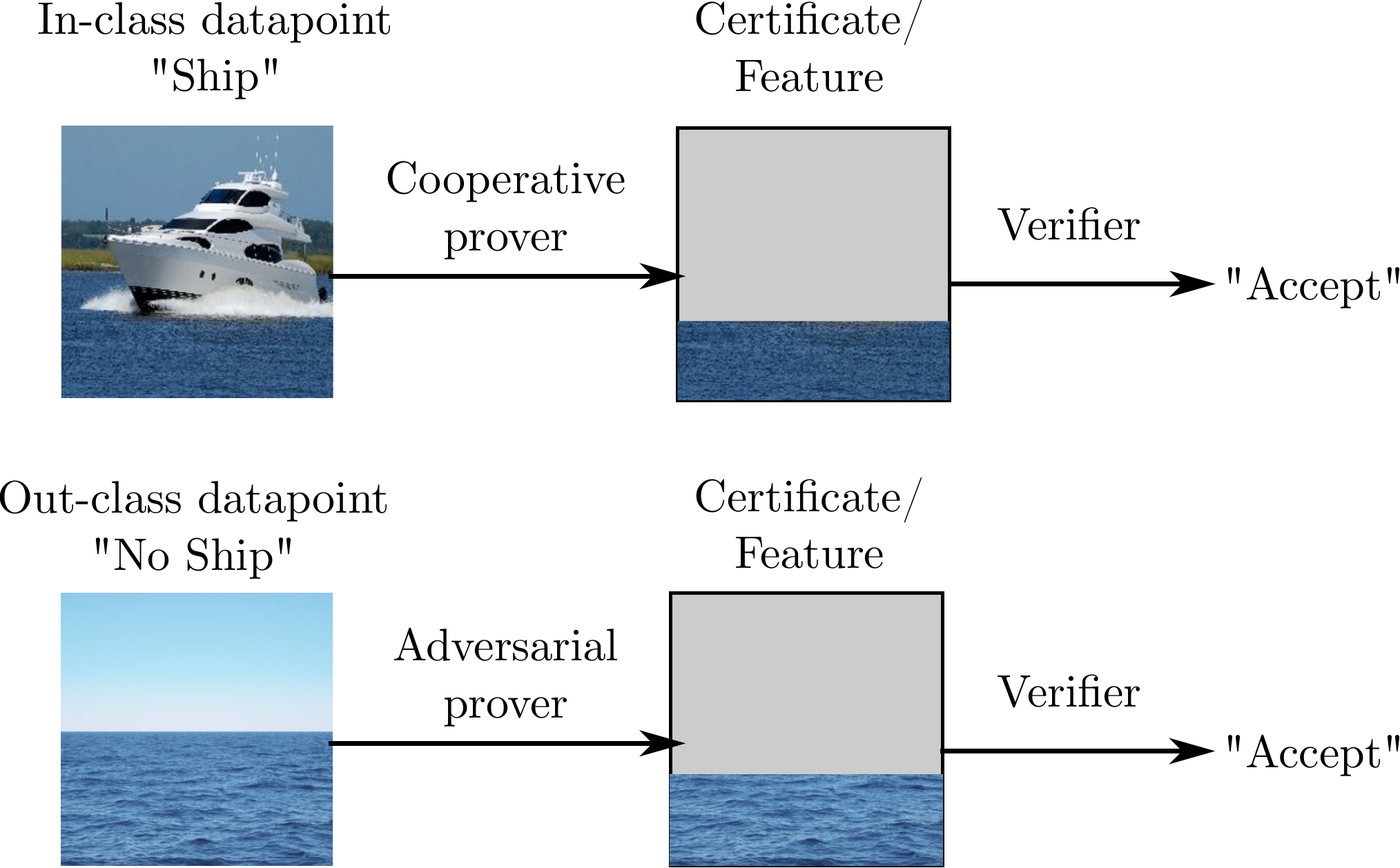}  &  \includegraphics[width=0.48\textwidth]{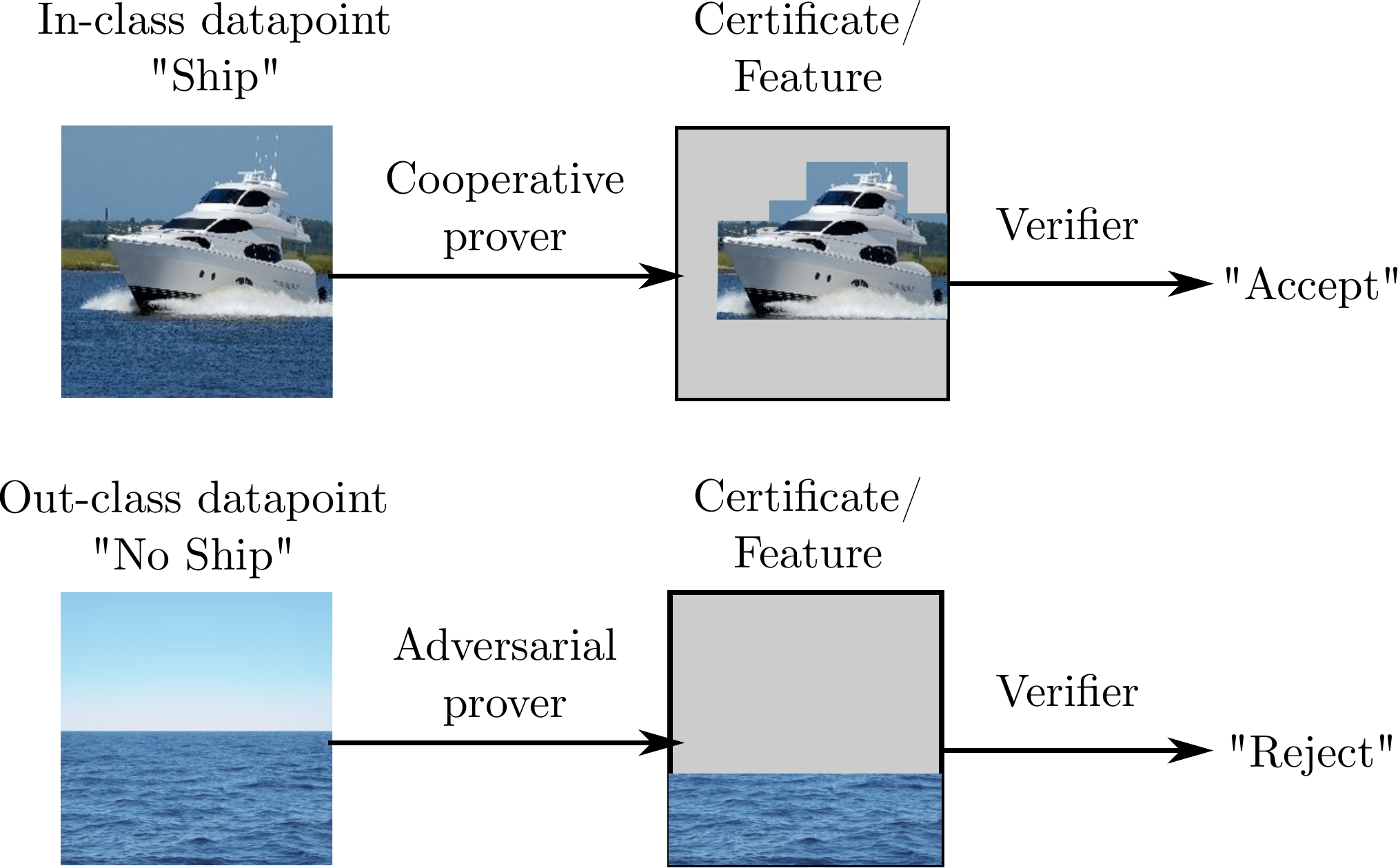}\\[1em]
        a) & b)
    \end{tabular}
    
    \caption{Illustration of interactive classification with an unsound and a sound verifier. a) The verifier Arthur accepts "water" features as certificates for the "ship" class. While these allow for high completeness, as most ships are on water, the soundness is low since Arthur can be convinced that images without ships belong to the class. Note that this behaviour can actually be observed when classifiers are trained on datasets where boat and water highly correlate~\cite{lapuschkin2016analyzing}.
    b) Arthur uses the strategy to only accept "boat" features as certificates. This strategy is both complete and sound since an adversarial prover cannot produce these certificates. }
    \label{fig:boats}
\end{figure}

\subsection{Interactive Classification}

Interactive classification is a concept inspired by interactive proof systems applied to the task of deciding whether datapoints belong to a certain class. We refer to~\Cref{fig:boats} for an illustration. The framework has been put first put forward in~\cite{chang2019game}, after it was noted that without an adversarial aspect, prover and verifier can cooperate to exchange uninformative features~\cite{yu2019rethinking}. Specifically, the Merlin-Arthur protocol~\cite{babai1985trading}  has inspired the formulation of an equivalent Merlin-Arthur classifier in~\cite{waldchen2022merlin}.
The prover Merlin receives a datapoint and selects a certificate (usually a relevant feature of the datapoint) to send to the verifier Arthur, who either accepts this as evidence for the class or rejects.
The prover can be cooperative, thus trying to convince Arthur of the correct class, or adversarial, trying to get Arthur to accept a datapoint outside the class.
The probability that the cooperative prover to convince Arthur for a random in-class datapoint is called the \emph{completeness} of the protocol. The probability that Arthur cannot be convinced for a random out-class datapoint is the \emph{soundness}.

The idea is that a commercial classifier could be mandated to be in interactive classifier with high completeness and soundness. An auditor would establish the soundness themselves by trying to fool Arthur to accept out-class datapoints. Contrary to the precision, completeness and soundness of the Merlin-Arthur pair can be readily estimated on the test dataset. The question is then whether completeness and soundness assure that highly-informative certificates/features are exchanged that can then be further examined for sensitive properties (e.g. sex or race in hiring decisions). In \cite{waldchen2022merlin}, the authors connect the soundness and completeness of the setup to a lower-bound on the average precision of the certificates exchanged between Merlin and Arthur. A crucial quantity that appears in this bound is the Asymmetric Feature Correlation.

\subsection{Asymmetric Feature Correlation}

Asymmetric Feature Correlation (AFC) was introduced in~\cite{waldchen2022merlin} as a possible quirk of the dataset where features that are individually not informative of the class can indicate the class via their correlation. That means, every feature appears equally likely inside and outside of the class, but the set of features can be highly correlated outside and highly anti-correlated inside. We give an illustration in~\Cref{fig:overview} b) and define it formally in~\Cref{def:AFC}. Merlin and Arthur can use these features to communicate the class. Despite using features with low precision, they retain high completeness since the features are spread out over the in-class datapoints and high soundness since Arthur only fails on the very few out-class datapoints where the features are concentrated. 
This means that AFC is a crucial quantity that connects high completeness and soundness to high informativeness of the features. We thus require an estimate or at least an upper bound of the AFC of the dataset.

The authors in~\cite{waldchen2022merlin} show that the AFC can be bounded by the maximum number of features in a datapoint. However, if one considers arbitrary subsets of the input as possible features, then this bound becomes exponentially large and thus unusable. One possible remedy is to further exploit the structure of the dataset to get a tighter bound on the AFC. In this work we provide a different argument: Finding such a feature set that realises a large AFC, even if it exists in principle, is closely related to finding highly-dense subgraphs in graphs, which is what we use to argue that this is a computationally hard task and cannot easily be exploited. This was also conjectured in~\cite{waldchen2022merlin}.

\section{Graph Theoretic formulation}

\begin{figure}
    \centering
    \begin{tabular}{c@{\hspace{1cm}}c}
         \includegraphics[width=0.4\textwidth]{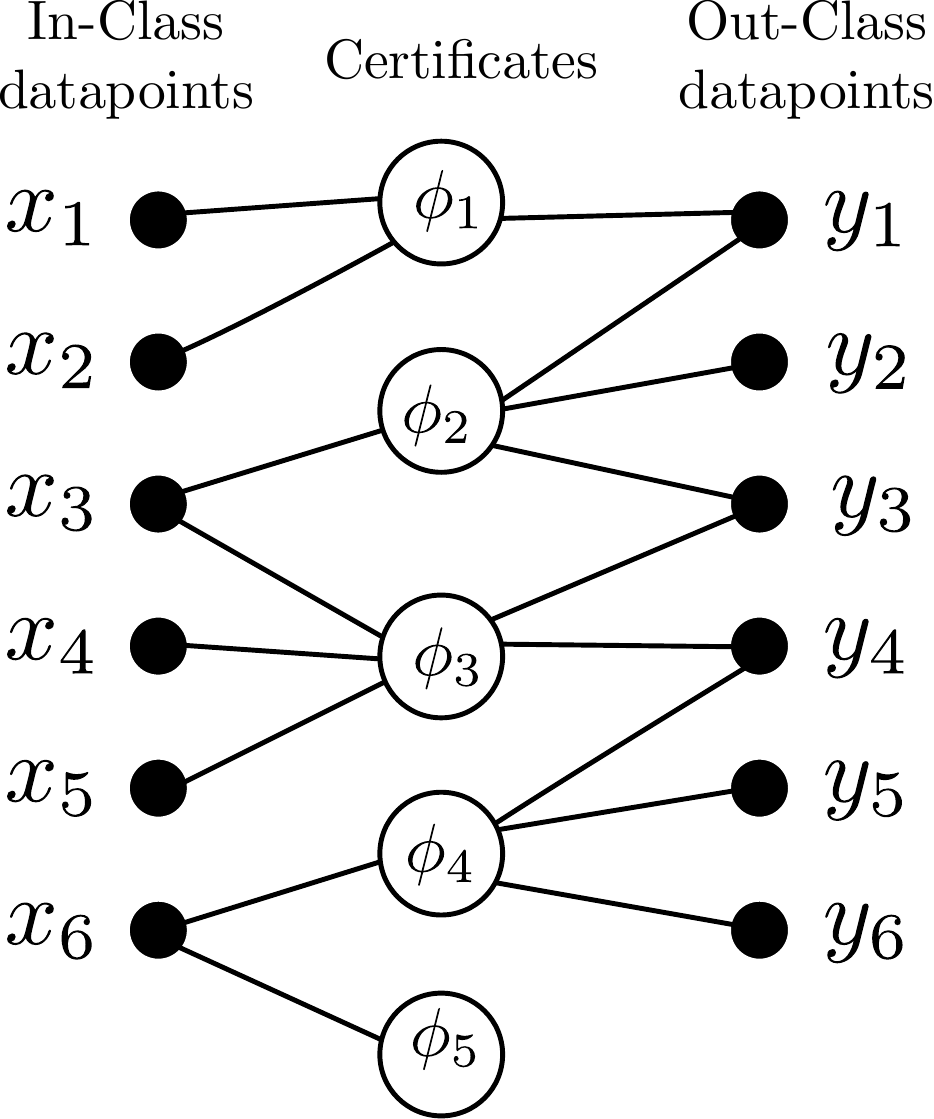} 
         & \includegraphics[width=0.35\textwidth]{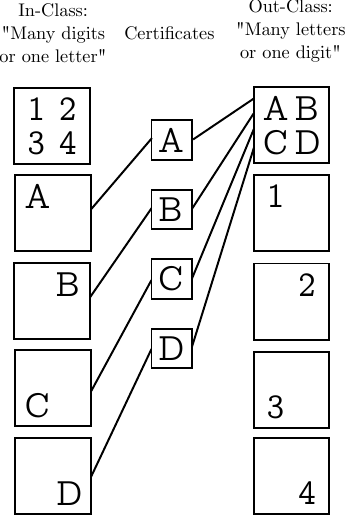} \\[1em]
       a)  & b)
    \end{tabular}
    
    \caption{\textbf{a)} An example instance of the Certificate-Selection problem as a tripartite graph $G=(V,E)$, where $V = D_{1} \cup C \cup D_{-1}$ and $E \subset (D_{1}\times C) \cup (D_{-1} \times C)$. A datapoint is connected to a certificate when it is possible to produce that certificate from this datapoint. An example would be text data, where the certificates are certain words that are indicative of the class. \textbf{b)} An example image dataset with an AFC of 4. The features signifying a letter have low precision (conditioned on ``A'', in-class and out-class are equally likely). 
    Yet, using letters as certificates for the class, Arthur can classify almost perfectly. He only fails for the top two images, where he cannot be convinced that the image with many digits is in the class, but accepts the image with many letters. Thus, the strategy achieves completeness and soundness of $\frac{4}{5}$, while using features with a low precision of 0.5. It is easy to see that this example can be expanded by including more letters and digits to a case of arbitrarily high soundness and completeness with uninformative features. This also illustrates why the AFC can be upper-bounded by the maximum number of features per datapoint.}
    \label{fig:overview}
\end{figure}

What reasonably constitutes a useful certificate is subject to ongoing debate. Most common are features in the form of subsets of the input vector, such as cutouts from an image, but others have proposed queries~\cite{chen2018learning}, anchors~\cite{ribeiro2018anchors} or more abstract functions\cite{anil2021learning}.
For our purposes, we do not need to specify the exact kind of certificate we consider. Rather, we can describe the relationship between datapoints and certificates as that of vertices in a graph. When a certificate can be produced for a datapoint, they share an edge in the graph. An illustration is given in~\Cref{fig:overview} a).

\begin{definition}[Certificate-Selection Instance (CSI)]
A certificate-selection instance is a tripartite graph $G=(V,E)$, where $V = D_{1} \cup C \cup D_{-1}$ and $E \subset (D_{1}\times C) \cup (D_{-1} \times C)$. We call the vertices in $D_1$ the in-class datapoints, the ones in $D_{-1}$ the out-class datapoints and the ones in $C$ the certificates.    
\end{definition}

We can now give the definition for the cooperative prover (Merlin) in our setup as an certificate assignment for each in-class datapoint. We do not explicitly consider the strategy of the adversarial prover which is assumed to play optimally and for every out-class datapoint selects a certificate that convinces Arthur if he is able to do so.
\begin{definition}[Cooperative Prover]
Given an CSI $G=(D_{1} \cup C \cup D_{-1}, E)$, a cooperative prover is an assignment $M: D_1 \rightarrow C$, such that $\forall x \in D_1: (x, M(x)) \in E$. The space of all provers of $G$ is $\mathcal{M}(G)$.
\end{definition}
The verifier (Arthur) has the task of deciding to accept a certificate for the class.
\begin{definition}[Verifier]
Given an CSI $G=(D_{1} \cup C \cup D_{-1}, E)$, a verifier is an assignment $A: C \rightarrow \skl{-1,1}$, corresponding to reject/accept of the certificate as evidence for the class. The space of all possible verifiers for G is $\mathcal{A}(G)$.
\end{definition}

We can then define soundness, completeness and the average precision for this setup.
The completeness is the probability that Merlin can convince Arthur of the class for the in-class datapoints:
\[
\compl(A,M) := \frac{\bkl{\skl{x \in D_1~|~ A(M(x)) = 1}}}{\bkl{D_1}}.
\]
The soundness is the probability that an out-class datapoint does not allow for a certificate that convinces Arthur of the class:
\[
\sound(A) := \frac{\bkl{\skl{x \in D_{-1}~|~ \forall \phi \in N(x): A(\phi) = -1}}}{\bkl{D_{-1}}},
\]
where $N(x)$ is the neighbourhood of $x$, so all certificates connected to this datapoint.
We can then define the notion of an \textsc{$(\epsilon_c,\epsilon_s)$-Certificate-Selection Instance}, as a CSI for which there exist prover $M$ and verifier $A$ that achieve \\$\compl(A,M) \geq 1- \epsilon_c$ and $\sound(A) \geq 1 - \epsilon_s$.

The \emph{precision} of a certificate is defined as:
\[
\PR(\phi) := \frac{\bkl{D_1 \cap N(\phi)}}{N(\phi)}.
\]
The precision of a feature set $F$ is then $\PR(F) := \frac{\bkl{D_1 \cap N(F)}}{N(F)}$,
where $N(F) = \bigcup_{\phi \in F} N(\phi)$. We can then deduce for the precision of the set of certificates accepted by Arthur that
\[
 \PR(A) := \PR(A^{-1}(1)) = 1 - \frac{\hat\epsilon_s}{1-\hat\epsilon_c + \hat\epsilon_s},
\]
where $\hat\epsilon_c = 1 - \max_M \compl(A,M)$ and  $\hat\epsilon_s = 1 - \max_M \sound(A,M)$.
The average precision of Merlin is defined as
\[
\PR(M) := \frac{1}{\bkl{D_1}} \sum_{x \in D_1} \PR(M(x)).
\]
Note that we only define these quantities wrt. to one class, whereas in~\cite{waldchen2022merlin}, these definitions always consider two-classes by maximising or minimising over the two classes.

The general idea is that we would like to draw a conclusion from the precision of Arthur $\PR(A)$, which can be observed in the interactive classification setup by measuring completeness and soundness, to the unobservable average precision of Merlin $\PR(M)$. The case of $\PR(A)\approx 1$ corresponds to a complete and sound protocol, whereas $\PR(M)\approx 1$ corresponds to the use of informative features. 
The question is thus: How small can $\PR(M)$ be made if Arthur and Merlin cooperate to exploit the AFC to still ensure high completeness and soundness?

We now define the graph theoretic version of the asymmetric feature correlation derived in~\cite{waldchen2022merlin}.
\begin{definition}[Asymmetric Feature Correlation (AFC)] \label{def:AFC}
For a CSI $G=(D_{1} \cup C \cup D_{-1}, E)$, the asymmtric feature correlation is defined as
\[
 \kappa := \max_{F \subset C}\frac{1}{\bkl{F_1^\ast}} \sum_{y \in F_1^\ast} \max_{\phi \in N(y)\cap F} \kappa(\phi, F)
\]
where
\[
 \kappa(\phi, F) = \frac{\bkl{N(\phi)\cap D_{-1}}\bkl{N(F)\cap D_{1}}}{\bkl{N(F)\cap D_{-1}}\bkl{N(\phi)\cap D_{1}}} \quad \text{and} \quad F_1^\ast = N(F) \cap D_1.
\]
\end{definition}
To make the intuition clearer: Given a set of certificates $F$, the quantity $\frac{\bkl{N(\phi)\cap D_{l}}}{\bkl{N(F)\cap D_{l}}}$ is a measure of how correlated the certificates are in class $l$. If they are all connected to the same datapoint this quantity is close to 1. If they share no datapoints, this quantity becomes minimally $\frac{1}{\bkl{F}}$ on average. Thus $\kappa(\phi, F)$ determines the ratio of this correlation between out-class and in-class. 
The rest of the expression corresponds to an expectation value over a distribution of certificates. This distribution results from taking an in-class datapoint $y$ uniformly at random from the set of all in-class datapoints that are connected to certificates in $F$, so $N(F) \cap D_1$ and then selecting the worst-case certificate connected to this feature. It is straight-forward to check that the example given in~\Cref{fig:overview} b) has an AFC of 4.






\subsection{Deceptive Feature Selection}

Let us assume that Merlin and Arthur collude to use uninformative features (low precision) while achieving acceptable values for completeness and soundness. According to~\cite{waldchen2022merlin}, the only way to achieve this is through exploiting the AFC of the dataset. We now want to investigate the computational difficulty of this task.

\begin{definition}[Deceptive-Certificate-Selection (DCS)]
 We define the \textsc{Deceptive-Certificate-Selection} problem as follows:
  \begin{description}
    \item[Given:] Two constants $\epsilon_c, \epsilon_s \geq 0$ and an instance $G$ of $(\epsilon_c, \epsilon_s)$-CSI.
    \item[Task:] Maximise $1-\PR(M)$ over $M\in \mathcal{M}(G)$ s.t. there exist $A\in \mathcal{A}(G)$ with $\compl(A,M)\geq 1 - \epsilon_c$ and $\sound(A) \geq 1 - \epsilon_s$.
  \end{description}
\end{definition}

When we can show that this problem is computationally hard to solve, we can argue that the exploitation of AFC will not be a problem big enough to prevent the use of interactive classification to find certificates with high precision.

\section{Inapproximability Bounds}

For our inapproximability bound, we make use of inapproximability of \textsc{Densest-$k$-subgraph} problem (D$k$S), which is conditional on the \textsc{Dense-vs-Random conjecture}. This conjecture has been established in \cite{bhaskara2010detecting} and has proven instrumental in deriving a series of inapproximability results, e.g. for \textsc{Minimum-$k$-Union}~\cite{chlamtavc2017minimizing} and \textsc{Red-Blue-Set Cover}~\cite{chlamtavc2023approximating}. 

\begin{theorem}\label{thm:DCSP}
 Assuming $\SP\neq\SNP$, there is no polynomial-time algorithm that solves the decision version of \textsc{Deceptive-Certificate-Selection}. Assuming furthermore the \textsc{Dense-vs-Random} Conjecture, then for every $\epsilon > 0$ there is no polynomial-time algorithm that achieves a better factor than $\mathcal{O}(m^{1/8-\epsilon})$ for the DCS, where $m$ is the number of certificates.
\end{theorem}

\begin{figure}
    \centering
    \begin{tabular}{c@{\hspace{2cm}}c}
        \includegraphics[width=0.3\textwidth]{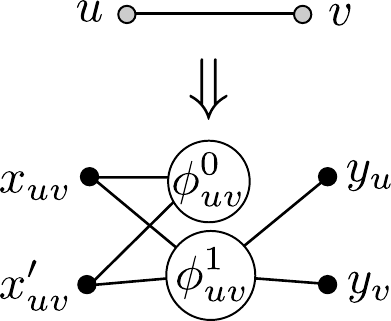}  &   \includegraphics[width=0.3\textwidth]{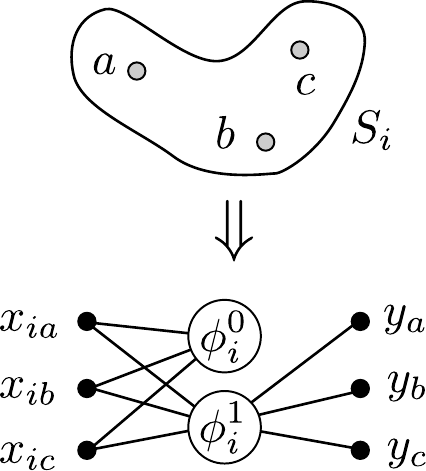}\\[1em]
         a) & b) 
    \end{tabular}
   
    \caption{a) Illustration of the reduction of \textsc{Densest-$k$-Subgraph} to \textsc{Deceptive-Feature-Selection}. For each vertex in the former, we add an out-class datapoint, and for each edge, we add two certificates and two in-class datapoints.
    b) Illustration of the reduction of \textsc{Min-$k$-Union} (of degree $r$) to \textsc{Deceptive-Feature-Selection-2}. For each element in the former, we add an out-class datapoint, and for each subset $S_i$ we add two certificates and $r$ in-class datapoints. }
    \label{fig:construction}
\end{figure}

\begin{proof}
We prove~\Cref{thm:DCSP} by showing that an $\alpha$-approximation of DCS implies an $\alpha^2$-approximation Densest-$k$-Subgraph, which cannot be approximated better than $\mathcal{O}(|V|^{1/4-\epsilon})$ assuming the Dense-vs-Random Conjecture~\cite{bhaskara2010detecting}.

Assume we are given a densest-$k$-subgraph instance $(G=(V,E), k)$. Let $\epsilon_c = \frac{1}{2|E|+1}$ and $\epsilon_s = \frac{k}{|V|}$. Then, we construct an instance of
$(\epsilon_c,\epsilon_s)-CSP$ in the following way:
For every vertex in $V$, we add a corresponding vertex in $D_{-1}$, so
\[
 D_{-1} = \bigcup_{v \in V } \skl{y_v}.
\]
For every edge in $E$, we add two datapoints in $D_1$, so
\[
 D_1 = \bigcup_{(u, v) \in E } \skl{x_{uv}, x^\prime_{uv}},
\]
and two certificates, one connected solely to the two datapoints in $x_{uv}$ and $x^\prime_{uv}$ and one connected to $x_{uv}, x^\prime_{uv}, y_u$ and $y_v$, thus
\[
 C = \bigcup_{(u, v) \in E } \skl{\phi_{uv}^0,\phi^1_{uv} } \quad\text{where}\quad 
  \begin{array}{c}
 \phi^0_{uv}=(x_{uv}, x^\prime_{uv}),\\[0.5em]  \phi^1_{uv} =(x_{uv}, x^\prime_{uv}, y_u, y_v).
 \end{array}
\]
\Cref{fig:construction} a) provides an illustration. It is easy to see that this instance can be classified with perfect soundness and completeness by just always exchanging the $\phi^0_{uv}$.

But to minimise the precision, Merlin wants to choose the uninformative features $\phi^1_{uv}$. Since we have chosen $\epsilon_c$ smaller than the number of edges, Arthur needs to accept all the certificates assigned by Merlin. Thus, Merlin's certificates can at most cover $k$ datapoints in $D_{-1}$, for Arthur to fulfil his soundness criterion.
The number of edges in the densest-$k$-subgraph thus determines how often Merlin can assign $\phi^1_{uv}$ instead of $\phi^0_{uv}$. 

For Merlin, it makes only sense to assign the same certificate to $x_{uv}$ and $x^\prime_{uv}$. If he chooses a $\phi^1_{uv}$ for $x_{uv}$, which Arthur needs to accept, this already adds $y_u$ and $y_v$ to the datapoints on which Arthur can be fooled.
Thus, choosing $\phi^1_{uv}$ for $x^\prime_{uv}$ reduces the precision with no cost on the soundness.
The precision of $\phi^0_{uv}$ and $\phi^1_{uv}$ is $1$ and $\frac{1}{2}$ respectively. If Merlin chooses $\phi^1$ $l$ times, we can calculate
\[
 1 - \PR(M) = \frac{l}{2}.
\]
Thus the $\SNP$-hardness of the decision version of CSP follows directly from the hardness of the decision version of D$k$S.

Now, let us assume that we had an approximation algorithm that achieves $1 - \PR(M) \geq \frac{OPT}{f(m)}$, where OPT is the maximum achievable and $f(m)$ is some function in the number of certificates. Then through our construction we would approximate the densest-$k$-subgraph with factor $f(m)$. Since the number of certificates in our construction is equal to twice the number of edges in the subgraph problem, and the number of edges is at most $|V|$, our algorithm guarantees a factor of $f(2|V|^2)$.
But since we know that D$k$S cannot be approximated better than $\mathcal{O}(n^{1/4 - \epsilon})$, we know that we cannot approximate DCSP better than $\mathcal{O}(n^{1/8 - \epsilon})$.\qed
\end{proof}

We can define a variant of the DCSP where instead of having low precision as target, we give it as a constraint and target high soundness as to not be detected by an auditor.

\begin{definition}[Deceptive-Certificate-Selection-2 Problem (DCS2)]
   We define the \textsc{Deceptive-Certificate-Selection-2} problem as follows.
  \begin{description}
    \item[Given:] An instance $G$ of $(\epsilon_c,\epsilon_s)-$CSP and constants $\epsilon_c, \epsilon_s, q \geq 0$.
    \item[Task:] Minimise $1-\sound(A)$ over $A\in \mathcal{A}(G)$ s.t. there exist $M\in \mathcal{M}(G)$ with $\compl(A,M)\geq 1-\epsilon_c$ and $\PR(M) \leq 1-q$.
  \end{description}
\end{definition}

For this problem, we can derive a conditional inapproximability factor of $\mathcal{O}(m^{1/4 - \epsilon)})$ via reduction of the \textsc{Min-$k$-Union} problem. The \textsc{Min-$k$-Union} considers a universe of elements $U$ and a selection of subsets $S\subset 2^{U}$, and searches for a collection of $k$ sets $S_i\in S$ that minimise $|\bigcup_{i=1}^k S_i|$. The inapproximability of the latter problem with a factor of $\mathcal{O}(m^{1/4 - \epsilon})$, where $m=|S|$, is  conditioned on the stronger assumption that extends the \textsc{Dense-vs-Random} conjecture to hypergraphs~\cite{chlamtavc2017minimizing}.

The reduction is analogous to our reduction for \textsc{Densest-$k$-Subgraph}. It is important to note that the reduction in~\cite{chlamtavc2017minimizing} considers $r$-uniform hypergraphs, and establishes the hardness of \textsc{Min-$k$-Union} also for the case where all sets $S_i$ have the same size $r=|S_i|$. Again, for every element in $U$ we add an out-class datapoint, and for every subset in $S$ with two certificates and $r$ in-class datapoints, as illustrated in~\Cref{fig:construction} b).
Then we set $\epsilon_c = \frac{1}{r|S|+1}$ and $q = \frac{k}{|S|}$. Again, Arthur must accept all of Merlin's feature to ensure the completeness criterion. To ensure low average precision, Merlin has to select at least $k$ $\phi^1$-certificates while trying to minimise the amount of out-class datapoints they cover. A solution to DCS2 thus solves \textsc{Min-$k$-Union} (of degree $r$). We do not get the $\frac{1}{2}$-exponent this time, because the inapproximability of \textsc{Min-$k$-Union} is directly stated in the number of sets.
Thus, we arrive at an inapproximability factor of $\mathcal{O}(m^{1/4 - \epsilon)}$ for DCS2, albeit conditional on the stronger \textsc{Hypergraph-Dense-vs-Random} conjecture.

\section{Discussion and Outlook}

These results must be interpreted carefully, as they do not yet imply that deceptive certificate selection cannot happen.
For a given CSI and given completeness and soundness, let $1-\PR^\ast$ be the solution of the DCS, in other words $\PR^\ast$ is the lowest possible average precision of the exchanged features.
We should expect the average precision of Merlin to lie between the precision of Arthur and the worst case, i.e.
\[
 1-\PR(A) \leq 1-\PR(M) \leq 1-\PR^\ast.
\]
So far, our analysis shows that it is computationally hard for $1-\PR(M)$ to approximate $1-\PR^\ast$. However, in the worst case $\frac{1-\PR^\ast}{1-\PR(A)}$ is of the order of $m$, the total numbers of certificates, as illustrated in~\Cref{fig:overview} b). An inapproximability result of $m^{1/8}$ does thus not yet guarantee that $1-\PR(M)$ will be close to $1-\PR(A)$, the quantity we can practically observe.
To ensure that $\PR(M)\approx \PR(A)$ we need either a stronger inapproximability with respect to $\PR^\ast$, or ideally directly show that certificate selection such that $1-\PR(M) \gg 1-\PR(A)$ is a computationally hard problem.

On the other hand, for practical purposes it is not necessarily required that $1-\PR(M)$ will be close to $1-\PR(A)$, only that $1-\PR(M) \ll 0.5$ as $0.5$ corresponds to uninformative features. 
Thus, a strong hardness-based separation from $1-\PR^\ast$ will be useful, since $1-\PR^\ast$ is upper-bounded by 1. At this point, we cannot say how informative Merlin's features are, but we can say that they will be far from uninformative. 

Furthermore, we want to stress that we are considering worst-case hardness, which does not necessarily imply that deceptive certificate selection will be hard on average. It can be easily shown that the classification problem itself is worst-case $\mathsf{NP}$-hard, yet existing machine-learning approaches successfully solve real-world instances. To make the argument stronger, future work might derive an average-case hardness bound. Of course, much depends on the distribution of the CSI-instances over which the average is taken, and thus it is important to investigate the graph structure of data and features for real-world datasets.

Reductions of the \textsc{Dense-vs-Random} conjecture hold promise in this direction. The conjecture is based on the hardness of finding dense subgraphs in random graphs, which makes the problem very suitable for average-case hardness reductions~\cite{chlamtavc2017minimizing}. Additionally, the log-density framework~\cite{bhaskara2010detecting} on which this conjecture is based, has been successfully used to design worst-case algorithms for a wide range of problems by studying algorithms for average-case instances~\cite{jones2023sum}. For \textsc{Minimum-$k$-Union} the approximation algorithm matches the lower provided by the conjecture~\cite{chlamtavc2017minimizing}. Ideally, one might analogously design an algorithm to approximate the solution of the DCS-problem and show that any result beyond that approximation factor is $\mathsf{NP}$-hard. This would give a solid understanding of how uninformative the certificates can be made with poly-time effort.

When extended in such a way, our results together with the work in~\cite{waldchen2022merlin} contribute to the establishment of a practically useful theoretical framework for explainable artificial intelligence (XAI). Notably, the interpretability guarantees do not make assumptions on the exact implementation of Arthur and Merlin and are thus applicable to neural network-based classifiers. This compares favourably to frameworks based on properties such as Lipschitz-constants or VC-dimensions, which do not yield useful results for parameter-rich neural networks~\cite{zhang2021understanding}. We thus expect the design of classifier setups for which interpretability guarantees can be proven, such as interactive classification, to be a fruitful approach for the field of formal interpretability and XAI in general.

Theoretical bounds derived with legible and reasonable assumptions are of great importance to make artificial intelligence-based systems trustworthy. Heuristic interpretability methods have been shown to fail as soon as an effort is made to obscure the reasoning of the classifier. With formal methods, on the other hand, we know under which conditions we can trust their results. Agents with high completeness and soundness could be mandated as commercial classifiers that allow for reliable auditing, e.g., in the context of hiring decisions. We can use past hiring decisions by the Merlin-Arthur classifier as ground truth, which implies perfect completeness. An auditor would use their own adversarial prover to check if Arthur has sufficient soundness. If so, the features by Merlin must be the basis of the hiring decisions. The auditor can then inspect them for protected attributes, e.g., race, sex or attributes that strongly correlate with them~\cite{mehrabi2021survey}.

Of course, certificates do not yet explain the whole reasoning process of the model. Crucially, they do not consider whether the certificate is causally related to the class or just correlated with it, which is an important consideration for features that are correlated with sensitive properties. However, there has been progress to adapt interactive classification to find causal features~\cite{chang2020invariant}.
Extending interpretability to higher-level reasoning, for example via multiple rounds of interaction between prover and verifier, is left for future research.

\section{Conclusion}

In this work, we abstract interactive classification into an optimisation problem on a tripartite graph.
We furthermore consider a deceptive prover-verifier duo that aims to use uninformative certificates and to exploit high asymmetric feature correlation to nevertheless achieve high completeness and soundness.  It turns out that this task is computationally hard (assuming $\SNP\neq \SP$), and hard to approximate (assuming the Dense-vs-Random conjecture).

When considering a machine learning task where the verifier will be trained on a training-dataset, then this deceptive strategy is additionally complicated by the learning problem. Soundness and completeness can be evaluated on a test dataset, and the strategy of choosing uninformative certificates developed during training needs to generalise.

These two barriers make it likely that even if the dataset has a high AFC, it will be difficult to exploit. That means, we can draw conclusions from the soundness and completeness of an interactive classification setup about the informativeness of the exchanged features.

%
\bibliographystyle{abbrvnat}
\bibliography{bibliography}

\begin{thebibliography}{28}
\providecommand{\natexlab}[1]{#1}
\providecommand{\url}[1]{\texttt{#1}}
\expandafter\ifx\csname urlstyle\endcsname\relax
  \providecommand{\doi}[1]{doi: #1}\else
  \providecommand{\doi}{doi: \begingroup \urlstyle{rm}\Url}\fi

\bibitem[Anders et~al.(2020)Anders, Pasliev, Dombrowski, M{\"u}ller, and
  Kessel]{anders2020fairwashing}
C.~Anders, P.~Pasliev, A.-K. Dombrowski, K.-R. M{\"u}ller, and P.~Kessel.
\newblock Fairwashing explanations with off-manifold detergent.
\newblock In \emph{International Conference on Machine Learning}, pages
  314--323. PMLR, 2020.

\bibitem[Anil et~al.(2021)Anil, Zhang, Wu, and Grosse]{anil2021learning}
C.~Anil, G.~Zhang, Y.~Wu, and R.~Grosse.
\newblock Learning to give checkable answers with prover-verifier games.
\newblock \emph{arXiv preprint arXiv:2108.12099}, 2021.

\bibitem[Babai(1985)]{babai1985trading}
L.~Babai.
\newblock Trading group theory for randomness.
\newblock In \emph{Proceedings of the seventeenth annual ACM symposium on
  Theory of computing}, pages 421--429, 1985.

\bibitem[Bhaskara et~al.(2010)Bhaskara, Charikar, Chlamtac, Feige, and
  Vijayaraghavan]{bhaskara2010detecting}
A.~Bhaskara, M.~Charikar, E.~Chlamtac, U.~Feige, and A.~Vijayaraghavan.
\newblock Detecting high log-densities: an o (n $1/4$) approximation for
  densest k-subgraph.
\newblock In \emph{Proceedings of the forty-second ACM symposium on Theory of
  computing}, pages 201--210, 2010.

\bibitem[Chang et~al.(2019)Chang, Zhang, Yu, and Jaakkola]{chang2019game}
S.~Chang, Y.~Zhang, M.~Yu, and T.~Jaakkola.
\newblock A game theoretic approach to class-wise selective rationalization.
\newblock \emph{Advances in neural information processing systems}, 32, 2019.

\bibitem[Chang et~al.(2020)Chang, Zhang, Yu, and Jaakkola]{chang2020invariant}
S.~Chang, Y.~Zhang, M.~Yu, and T.~Jaakkola.
\newblock Invariant rationalization.
\newblock In \emph{International Conference on Machine Learning}, pages
  1448--1458. PMLR, 2020.

\bibitem[Chattopadhyay et~al.(2022)Chattopadhyay, Slocum, Haeffele, Vidal, and
  Geman]{chattopadhyay2022interpretable}
A.~Chattopadhyay, S.~Slocum, B.~D. Haeffele, R.~Vidal, and D.~Geman.
\newblock Interpretable by design: Learning predictors by composing
  interpretable queries.
\newblock \emph{arXiv preprint arXiv:2207.00938}, 2022.

\bibitem[Chen et~al.(2018)Chen, Song, Wainwright, and Jordan]{chen2018learning}
J.~Chen, L.~Song, M.~Wainwright, and M.~Jordan.
\newblock Learning to explain: An information-theoretic perspective on model
  interpretation.
\newblock In \emph{International Conference on Machine Learning}, pages
  883--892. PMLR, 2018.

\bibitem[Chlamt{\'a}{\v{c}} et~al.(2017)Chlamt{\'a}{\v{c}}, Dinitz, and
  Makarychev]{chlamtavc2017minimizing}
E.~Chlamt{\'a}{\v{c}}, M.~Dinitz, and Y.~Makarychev.
\newblock Minimizing the union: Tight approximations for small set bipartite
  vertex expansion.
\newblock In \emph{Proceedings of the Twenty-Eighth Annual ACM-SIAM Symposium
  on Discrete Algorithms}, pages 881--899. SIAM, 2017.

\bibitem[Chlamt{\'a}{\v{c}} et~al.(2023)Chlamt{\'a}{\v{c}}, Makarychev, and
  Vakilian]{chlamtavc2023approximating}
E.~Chlamt{\'a}{\v{c}}, Y.~Makarychev, and A.~Vakilian.
\newblock Approximating red-blue set cover.
\newblock \emph{arXiv preprint arXiv:2302.00213}, 2023.

\bibitem[Dimanov et~al.(2020)Dimanov, Bhatt, Jamnik, and
  Weller]{dimanov2020you}
B.~Dimanov, U.~Bhatt, M.~Jamnik, and A.~Weller.
\newblock You shouldn't trust me: Learning models which conceal unfairness from
  multiple explanation methods.
\newblock In \emph{SafeAI@ AAAI}, 2020.

\bibitem[Holzinger et~al.(2017)Holzinger, Biemann, Pattichis, and
  Kell]{holzinger2017we}
A.~Holzinger, C.~Biemann, C.~S. Pattichis, and D.~B. Kell.
\newblock What do we need to build explainable ai systems for the medical
  domain?
\newblock \emph{arXiv preprint arXiv:1712.09923}, 2017.

\bibitem[Jones et~al.(2023)Jones, Potechin, Rajendran, and Xu]{jones2023sum}
C.~Jones, A.~Potechin, G.~Rajendran, and J.~Xu.
\newblock Sum-of-squares lower bounds for densest $ k $-subgraph.
\newblock \emph{arXiv preprint arXiv:2303.17506}, 2023.

\bibitem[Lapuschkin et~al.(2016)Lapuschkin, Binder, Montavon, Muller, and
  Samek]{lapuschkin2016analyzing}
S.~Lapuschkin, A.~Binder, G.~Montavon, K.-R. Muller, and W.~Samek.
\newblock Analyzing classifiers: Fisher vectors and deep neural networks.
\newblock In \emph{Proceedings of the IEEE Conference on Computer Vision and
  Pattern Recognition}, pages 2912--2920, 2016.

\bibitem[Lapuschkin et~al.(2019)Lapuschkin, W{\"a}ldchen, Binder, Montavon,
  Samek, and M{\"u}ller]{lapuschkin2019unmasking}
S.~Lapuschkin, S.~W{\"a}ldchen, A.~Binder, G.~Montavon, W.~Samek, and K.-R.
  M{\"u}ller.
\newblock Unmasking clever hans predictors and assessing what machines really
  learn.
\newblock \emph{Nature communications}, 10\penalty0 (1):\penalty0 1--8, 2019.

\bibitem[Lundberg and Lee(2017)]{NIPS2017_7062shap}
S.~M. Lundberg and S.-I. Lee.
\newblock A unified approach to interpreting model predictions.
\newblock In I.~Guyon, U.~V. Luxburg, S.~Bengio, H.~Wallach, R.~Fergus,
  S.~Vishwanathan, and R.~Garnett, editors, \emph{Advances in Neural
  Information Processing Systems 30}, pages 4765--4774. Curran Associates,
  Inc., 2017.

\bibitem[Macdonald et~al.(2019)Macdonald, W{\"a}ldchen, Hauch, and
  Kutyniok]{macdonald2019rate}
J.~Macdonald, S.~W{\"a}ldchen, S.~Hauch, and G.~Kutyniok.
\newblock A rate-distortion framework for explaining neural network decisions.
\newblock \emph{arXiv preprint arXiv:1905.11092}, 2019.

\bibitem[Mehrabi et~al.(2021)Mehrabi, Morstatter, Saxena, Lerman, and
  Galstyan]{mehrabi2021survey}
N.~Mehrabi, F.~Morstatter, N.~Saxena, K.~Lerman, and A.~Galstyan.
\newblock A survey on bias and fairness in machine learning.
\newblock \emph{ACM Computing Surveys (CSUR)}, 54\penalty0 (6):\penalty0 1--35,
  2021.

\bibitem[Mohseni et~al.(2021)Mohseni, Zarei, and
  Ragan]{mohseni2021multidisciplinary}
S.~Mohseni, N.~Zarei, and E.~D. Ragan.
\newblock A multidisciplinary survey and framework for design and evaluation of
  explainable ai systems.
\newblock \emph{ACM Transactions on Interactive Intelligent Systems (TiiS)},
  11\penalty0 (3-4):\penalty0 1--45, 2021.

\bibitem[Ribeiro et~al.(2018)Ribeiro, Singh, and Guestrin]{ribeiro2018anchors}
M.~T. Ribeiro, S.~Singh, and C.~Guestrin.
\newblock Anchors: High-precision model-agnostic explanations.
\newblock In \emph{Proceedings of the AAAI conference on artificial
  intelligence}, volume~32, 2018.

\bibitem[Schraagen et~al.(2020)Schraagen, Elsasser, Fricke, Hof, and
  Ragalmuto]{schraagen2020trusting}
J.~M. Schraagen, P.~Elsasser, H.~Fricke, M.~Hof, and F.~Ragalmuto.
\newblock Trusting the x in xai: Effects of different types of explanations by
  a self-driving car on trust, explanation satisfaction and mental models.
\newblock In \emph{Proceedings of the Human Factors and Ergonomics Society
  Annual Meeting}, volume~64, pages 339--343. SAGE Publications Sage CA: Los
  Angeles, CA, 2020.

\bibitem[Slack et~al.(2020)Slack, Hilgard, Jia, Singh, and
  Lakkaraju]{slack2020fooling}
D.~Slack, S.~Hilgard, E.~Jia, S.~Singh, and H.~Lakkaraju.
\newblock Fooling lime and shap: Adversarial attacks on post hoc explanation
  methods.
\newblock In \emph{Proceedings of the AAAI/ACM Conference on AI, Ethics, and
  Society}, pages 180--186, 2020.

\bibitem[Slack et~al.(2021)Slack, Hilgard, Lakkaraju, and
  Singh]{slack2021counterfactual}
D.~Slack, A.~Hilgard, H.~Lakkaraju, and S.~Singh.
\newblock Counterfactual explanations can be manipulated.
\newblock \emph{Advances in neural information processing systems},
  34:\penalty0 62--75, 2021.

\bibitem[Vergara and Est{\'e}vez(2014)]{vergara2014review}
J.~R. Vergara and P.~A. Est{\'e}vez.
\newblock A review of feature selection methods based on mutual information.
\newblock \emph{Neural computing and applications}, 24:\penalty0 175--186,
  2014.

\bibitem[W{\"a}ldchen et~al.(2022{\natexlab{a}})W{\"a}ldchen, Pokutta, and
  Huber]{waldchen2022training}
S.~W{\"a}ldchen, S.~Pokutta, and F.~Huber.
\newblock Training characteristic functions with reinforcement learning:
  Xai-methods play connect four.
\newblock In \emph{International Conference on Machine Learning}, pages
  22457--22474. PMLR, 2022{\natexlab{a}}.

\bibitem[W{\"a}ldchen et~al.(2022{\natexlab{b}})W{\"a}ldchen, Sharma, Zimmer,
  and Pokutta]{waldchen2022merlin}
S.~W{\"a}ldchen, K.~Sharma, M.~Zimmer, and S.~Pokutta.
\newblock Formal interpretability with merlin-arthur classifiers.
\newblock \emph{arXiv preprint arXiv:2206.00759}, 2022{\natexlab{b}}.

\bibitem[Yu et~al.(2019)Yu, Chang, Zhang, and Jaakkola]{yu2019rethinking}
M.~Yu, S.~Chang, Y.~Zhang, and T.~Jaakkola.
\newblock Rethinking cooperative rationalization: Introspective extraction and
  complement control.
\newblock In \emph{Proceedings of the 2019 Conference on Empirical Methods in
  Natural Language Processing and the 9th International Joint Conference on
  Natural Language Processing (EMNLP-IJCNLP)}, pages 4094--4103. Association
  for Computational Linguistics, 2019.

\bibitem[Zhang et~al.(2021)Zhang, Bengio, Hardt, Recht, and
  Vinyals]{zhang2021understanding}
C.~Zhang, S.~Bengio, M.~Hardt, B.~Recht, and O.~Vinyals.
\newblock Understanding deep learning (still) requires rethinking
  generalization.
\newblock \emph{Communications of the ACM}, 64\penalty0 (3):\penalty0 107--115,
  2021.

\end{thebibliography}
%





\end{document}